# Inference in Polytrees with Sets of Probabilities

José Carlos Ferreira da Rocha[1,2]    Fabio Gagliardi Cozman[1]    Cassio Polpo de Campos[1,3]

[1]Escola Politécnica, Univ. de São Paulo, Av. Prof. Mello Moraes, 2231, São Paulo, SP - Brazil
[2]Univ. Estadual de Ponta Grossa, Ponta Grossa, PR - Brazil
[3]Pontifícia Universidade Católica, São Paulo, SP - Brazil
fgcozman@usp.br, jrocha@uepg.br, cassio@ime.usp.br

## Abstract

Inferences in directed acyclic graphs associated with probability intervals and sets of probabilities are NP-hard, *even* for polytrees. We propose: 1) an improvement on Tessem's A/R algorithm for inferences on polytrees associated with probability intervals; 2) a new algorithm for approximate inferences based on local search; 3) branch-and-bound algorithms that combine the previous techniques. The first two algorithms produce complementary approximate solutions, while branch-and-bound procedures can generate either exact or approximate solutions. We report improvements on existing techniques for inference with probability sets and intervals, in some cases reducing computational effort by several orders of magnitude.

## 1 Introduction

A *credal network* provides a representation for imprecise probabilistic statements through direct acyclic graphs [7, 10, 17]. Such graph-theoretical models can be viewed as Bayesian networks with relaxed numerical parameters: each node in the graph represents a random variable, and each variable is associated with a set of probability distributions. The structure of the graph indicates relations of independence between variables; the "size" of the sets of probabilities encodes the imprecision in probability values. Such a model can be used to study robustness of probabilistic models, to investigate the behavior of groups of experts, or to represent incomplete or vague knowledge about probabilities [27]. In Section 2 we review the concepts of credal sets and credal networks, and in Section 3 we review the concept of *strong independence* and discuss justifications for it — we also present a justification based on mutual information.

An *inference* with a credal network is the computation of upper and lower probabilities for each category of a *query* variable. This computation, under the most commonly adopted semantics for credal networks (using strong independence) is NP-hard even for polytrees [12]. Exact and approximate algorithms have been proposed in the literature, but no algorithm can handle large credal networks exactly — depending on the characteristics of the network, even networks with a few nodes can present unsurmountable difficulties.

In this paper we propose novel algorithms for marginal inference in polytrees. We present an extension of Tessem's A/R approximate algorithm [26]; we essentially combine the original A/R algorithm with the recently developed SVE algorithm [12]. Approximations with the new algorithm, called A/R+, are significantly better (often by an order of magnitude) than approximations produced by the original A/R algorithm. We then explore properties of fractional multilinear programming to produce a local search algorithm that is, in some sense, "complementary" to A/R+; we demonstrate that this new search algorithm is extremely efficient in finding excellent approximations. We then combine both algorithms using branch-and-bound methods that can produce either exact or approximate probability bounds. Finally, we present examples showing the gains obtained by these techniques.

## 2 Credal sets and credal networks

A convex set of probability distributions is called a *credal set* [21]. A credal set for $X$ is denoted by $K(X)$; we assume that every variable is categorical and that every credal has a finite number of vertices. A conditional credal set is a set of conditional distributions, obtained applying Bayes rule to each distribution in a credal set of joint distributions [27]. Given a number of marginal and conditional credal sets, an *extension* of these sets is a joint credal set with the given marginal and conditional credal sets. In this paper we are exclusively concerned with computing the largest possible extension for a collection of marginal and conditional credal sets.

Given a credal set $K(X)$ and a function $f(X)$, the

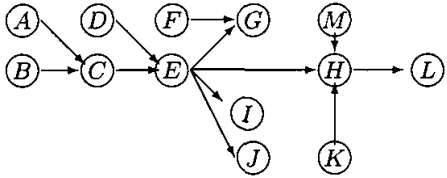

Figure 1: A polytree.

*lower* and *upper* expectations of $f(X)$ are defined respectively as $\underline{E}[f(X)] = \min_{p(X) \in K(X)} E_p[f(X)]$ and $\overline{E}[f(X)] = \max_{p(X) \in K(X)} E_p[f(X)]$ (here $E_p[f(X)]$ indicates standard expectation). Similarly, the *lower probability* and the *upper probability* of event $A$ are defined respectively as $\underline{P}(A) = \min_{p(X) \in K(X)} P(A)$ and $\overline{P}(A) = \max_{p(X) \in K(X)} P(A)$.

A *credal network* is a directed acyclic graph where each node of the graph is associated with a variable $X_i$ and with a collection of conditional credal sets $K(X_i|\text{pa}(X_i))$, where $\text{pa}(X_i)$ denotes the parents of $X_i$ in the graph (note that we have a conditional credal set for each value of $\text{pa}(X_i)$). A root node is associated with a single marginal credal set. Figure 1 shows the structure of a credal network that is later used in examples.

The sets $K(X|\text{pa}(X))$ are *separately specified* when there is no constraint on the conditional set $K(X|\text{pa}(X) = \pi_1)$ that is based on the properties of $K(X|\text{pa}(X) = \pi_2)$, for any $\pi_2 \neq \pi_1$ — that is, the conditional sets bear no relationship amongst them. In this paper we assume that local credal sets are always separately specified. Justifications for this separability assumption can be found in [12]; in essence, imprecision in probability values can be naturally expressed by separately specified sets, and non-separately specified sets face difficulties with the concept of conditional independence.

## 3 Strong independence, mutual information and strong extensions

We take that in a credal network every variable is independent of its nondescendants nonparents given its parents. Such an assumption obviously depends on the particular concept of independence adopted for a credal network, and in the theory of credal sets we find several distinct definitions of independence [8, 10, 14]. In this paper we adopt the concept of strong *independence*: two variables $X$ and $Y$ are strongly independent when every extreme point of $K(X,Y)$ satisfies standard stochastic independence of $X$ and $Y$ (that is, $p(X|Y) = p(X)$ and $p(Y|X) = p(Y)$) [10]. This definition is usually applied to convex sets of probabilities; note that only the extreme points of $K(X,Y)$ must satisfy stochastic independence.

Strong independence is the most commonly adopted concept of independence for credal sets, presumably because it is an obvious generalization of the standard concept of stochastic independence. The concept can also be derived from the decision-theoretic concept of *epistemic independence* and additional conditions of separability [11]. Here we propose an additional motivation for the concept of strong independence.

Suppose we have two credal sets $K(X)$ and $K(Y)$, and we want to form the set of joint distributions $p(X,Y)$ such that $X$ and $Y$ have null mutual information — a most reasonable condition for two variables to be termed "independent." We then obtain a *non-convex* joint set $Q(X,Y)$ that satisfies strong independence.[1] So, apart from lack of convexity, strong independence is a consequence of mutual information constraints, a very simple motivation for the concept. Readers who wish to use this mutual information argument and retain convexity can reason that a credal set and its convex hull produce the same lower/upper expectations, and therefore the convex hull of $Q(X,Y)$ can be used as the "correct" joint credal set.

The *strong extension* of a credal network is the largest joint credal set such that every variable is strongly independent of its nondescendants nonparents given its parents. The strong extension of a credal network is the joint credal set that contains every possible combination of vertices for all credal sets in the network [11]; that is, each vertex of a strong extension factorizes as follows:

$$p(X_1,\ldots,X_n) = \prod_i p(X_i|\text{pa}(X_i)). \quad (1)$$

## 4 Inference with strong extensions

A *marginal inference* in a credal network is the computation of lower/upper probabilities in an extension of the network. If $X_q$ is a *query* variable and $\mathbf{X}_E$ represents a set of *observed* variables, then an inference is the computation of tight bounds for $p(X_q|\mathbf{X}_E)$ for one or more values of $X_q$. For inferences in strong extensions, it is known that the distributions that minimize/maximize $p(X_q|\mathbf{X}_E)$ belong to the set of vertices of the extension [17].

The problem we face is one of combinatorial optimization; we must find a vertex for each local credal set $K(X_i|\text{pa}(X_i))$ so that Expression (1) leads to a maximum/minimum of $p(X_q|\mathbf{X}_E)$. Searching the possible vertices of the strong extension for these maxima/minima does not seem to be easy; the only known polynomial algorithm for strong extensions is the 2U algorithm, which processes polytrees with *binary* variables only [17]. Other than this "pocket" of tractability, all other situations seem to offer tremendous computational challenges. In general, inference is a NP-hard problem (even for polytrees) [12]. The

---
[1] The set $Q(X,Y)$ contains *only* distributions that satisfy strong independence. The failure of convexity is caused by the fact that the mixture of two distributions that satisfy independence may fail to display independence [25].

difficulty faced by inference algorithms is the potentially enormous number of vertices that a strong extension can have, even for small networks. Consider a network with three nodes,

$$(X) \rightarrow (Y) \leftarrow (Z), \qquad (2)$$

where $X$, $Y$ and $Z$ have four values each, and where all credal sets have four vertices each.[2] There are $4^{18}$ (about 69 billion) potential vertices of the strong extension (that is, $4^{18}$ different distributions satisfying factorization (1)). Existing exact algorithms cannot overcome such situations, and consequently face difficulties for every network that contains a fragment such as (2).[3]

The complexity of a credal network is a function of the structure of the network, the number of values for variables in the network, and the number of vertices of credal sets. For the polytree in Figure 1, if the variables are ternary and the credal sets have three extreme points, then we have $3^{69}$ (about $835 \times 10^{30}$) potential combinations of vertices that may be vertices of the strong extension.

Exact inference algorithms based on enumeration examine all potential vertices of the strong extension to produce the required lower/upper values [4, 7, 10]. Currently, the most efficient enumeration algorithm seems to be the Separable Variable Elimination (SVE) algorithm [12]. SVE explores properties of separately specified sets to reduce the number of vertices that must be examined (by keeping the sets in a separable form, and by applying convex hull operations to the separately specified sets). The SVE algorithm can be used in relatively large networks, however the algorithm can be easily dwarfed in seemingly trivial situations: for example, the network (2) would be an impossible task for SVE, given the need to handle billions of functions at a time.

The computation of lower/upper values for $p(X_q|\mathbf{X}_E)$ is the minimization/maximization of a fraction containing polynomials in probability values. In Section 6 we return to this idea in the context of approximate algorithms. We distinguish *outer* and *inner* approximations: the former produce intervals that enclose the correct probability interval between lower and upper probabilities, while the latter produce intervals that are enclosed by the correct probability interval. Outer approximations can be found in [6, 18, 26]; inner approximations can be found in [1, 5, 4, 9].

---

[2]It makes sense to consider that every credal set has at least as many vertices as the number of values of the associated variable, as the simplest model for credal sets is the $\epsilon$-contaminated class, which naturally leads to such a relationship between vertices and values [10].

[3]We remark that the branch-and-bound procedure to be described in Section 7 can easily handle situations such as (2).

## 5　Outer approximation: The A/R+ algorithm

The most straightforward scheme for outer approximations in credal polytrees is currently Tessem's A/R algorithm [26]. The first assumption in Tessem's algorithm is that every credal set is approximated by a collection of probability intervals. Such approximation is always possible (and always an outer approximation), as we can obtain the probability interval

$$\left[\min_{p(X) \in K(X)} p(x_j), \max_{p(X) \in K(X)} p(x_j)\right] \qquad (3)$$

for any value $x_j$ of a variable $X$ (and likewise for conditional probability values). Obviously the replacement of credal sets by probability intervals introduces potential inaccuracies into inferences.

Tessem's central idea was to generalize Pearl's belief propagation algorithm to accommodate probability intervals (in an approximate way). The functions $\lambda$ and $\pi$ used in belief propagation are still defined with identical purposes, but they are now interval-valued functions. These intervals are manipulated using interval arithmetic and two additional techniques called *annihilation* and *reinforcement* (thus the name A/R).

To understand the mechanics of A/R, it is interesting to look at a particular operation, the computation of the interval-valued function $\pi(X)$. This function is computed at a node $X$ with parents $Y_0, \ldots, Y_k$. Consider then the computation of $\pi_*(x_j)$, the lower bound of $\pi(x_j)$ for a particular value $x_j$:

1. Construct the interval-valued function $\beta(Y_0, \ldots, Y_k)$ by interval-multiplication of the messages $\pi_X(Y_i)$ received by $X$ (these messages are also interval-valued).

2. Construct a distribution $p(Y_0, \ldots, Y_k)$ that is consistent with the intervals in $\beta(Y_0, \ldots, Y_k)$, such that $p(Y_0, \ldots, Y_k)$ minimizes the sum $\sum_{Y_0, \ldots, Y_k} \underline{p}(x_j|Y_0, \ldots, Y_k) p(Y_0, \ldots, Y_k)$, where $\underline{p}(x_j|Y_0, \ldots, Y_k)$ is the lower value for $p(x_j|Y_0, \ldots, Y_k)$; the minimum of the sum is $\pi_*(x_j)$.

These operations are efficient because it is not hard to find $p(Y_0, \ldots, Y_k)$ in step 2: sort $\underline{p}(x_j|Y_0, \ldots, Y_k)$ in increasing order, and distribute probability mass (consistently with $\beta(Y_0, \ldots, Y_k)$) from the smallest to the largest value of $\underline{p}(x_j|Y_0, \ldots, Y_k)$. The same operations can be adapted to compute the upper bound $\pi^*(x_j)$.

The A/R algorithm prescribes similar operations for computation of messages $\lambda_X(Y_i)$ and $\pi_{Z_i}(X)$ (where $Z_i$ is a child of $X$). The function $\lambda(X)$ is obtained by direct interval multiplication. The algorithm uses annihilation or reinforcement operations to "normalize" the functions $\lambda_X(Y_i)$, $\pi_{Z_i}(X)$, and the product $\pi(X)\lambda(X)$ — "normalization" means simply computing bounds that account for the fact that probability distributions add up to one.

The A/R algorithm is clever, but it can be significantly improved as follows. Consider a new method to compute $\pi(X)$ in the message propagation scheme:

1. For each interval-valued message $\pi_X(Y_i)$ received by $X$, create a *credal set* $\pi_X(Y_i)$ that is the largest credal set with lower/upper probabilities represented by $\pi_X(Y_i)$. Such a credal set can be easily generated [20].

2. Eliminate each parent $Y_i$ by combining vertices of $\pi_X(Y_i)$ with vertices of $K(X|Y_1,\ldots,Y_k)$, applying a convex hull algorithm on each combined set to eliminate redundant vertices. For example, start by computing $\sum_{Y_1} p(X|Y_1,\ldots,Y_k)p(Y_1)$ for each combination of $\{Y_2,\ldots,Y_k\}$, and try to eliminate redundant vertices for each combination of $\{Y_2,\ldots,Y_k\}$ — it is possible to do so because the sets are assumed separately specified [12].

3. Use the resulting credal set $K(X)$ to produce probability intervals $\pi(X)$ through Expression (3).

The reader familiar with the Separable Variable Elimination (SVE) algorithm [12] will realize that step 2 of this procedure is exactly a "local" application of SVE. In a sense, the procedure just described is an "hybrid" of A/R and SVE: it uses SVE locally, collapsing potentially complex credal sets into probability intervals. The use of convex hull operations to eliminate redundancies is discussed in [12]; in practice we find that convex hull operations have a non-trivial cost and should be used carefully. One situation where such operations *must* be used is when several variables involved in the computation are binary, because the convex hull of points in one dimension can be easily computed. As the dimensionality of variables increases, the use of convex hull operations becomes less viable — experiments reported in [12] suggest that convex hull operations are not viable beyond four or five dimensions.

Denote the approximation of $\pi(x_j)$ generated in the previous paragraph by $\hat{\pi}_*$. Similar computations lead to an upper bound denoted by $\hat{\pi}^*$.

The operations that produce $\hat{\pi}_*$ can also be easily extended to other interval-valued messages used in belief propagation. The messages $\pi_{Z_i}(X)$ are computed using the same bounds of $\pi(X)$. The same basic procedure can be adapted to the computation of $\lambda_X(Y_i)$. We thus obtain the *A/R+* algorithm:
*run all the steps of the A/R algorithm, but whenever interval-valued $\pi_{Z_i}(X)$ messages must be multiplied, convert the messages into credal sets, run SVE locally, and convert the results back to probability intervals.*

Table 1: Mean relative error for inference $\overline{P}(E = e_0)$, for graph in Figure 1. Tests with three different combinations of number of categories and number of vertices; ensembles of 30, 15 and 5 sample networks respectively.

| number of categories | number of vertices | A/R mean error | A/R+ mean error |
|---|---|---|---|
| 03 | 02 | 0.38 | 0.03 |
| 03 | 03 | 0.15 | 0.01 |
| 04 | 02 | 0.27 | 0.04 |

The basic fact about A/R+ is that

**Theorem 1** *Interval-valued messages generated by the A/R+ algorithm are included or equal to interval-valued messages generated by the A/R algorithm, and include or are equal to probability intervals generated by the correct (set-valued) messages sent by SVE.*

*Proof.* Every message in A/R+ is generated by recursively taking a collection of messages used in the A/R algorithm and generating a more accurate representation for their product than the interval product; therefore A/R+ messages are enclosed by A/R messages. Also, A/R+ messages recursively contain all distributions that could possibly be generated in the correct propagation, as Expression (3) always starts the process with outer approximations. QED

More importantly, we have observed that outer approximations generated by the A/R+ algorithm are much more accurate than the ones produced by A/R. The apparently mild difference between the algorithms leads to an order of magnitude improvement in the bounds, as is illustrated by the following experiments.

First consider the network in Figure 1. We generated randomly 30 different credal networks with the same graph indicated in that figure. Variables have from 2 to 4 categories, and generated credal sets have 2 or 3 vertices; the vertices are generated so as to cover uniformly the space of distributions [19]. In Table 1 we show the mean relative errors in the computation of $\overline{P}(E = e_0)$, clearly showing that A/R+ reduces significantly the error (relative error is the absolute difference between approximate and correct values, divided by correct value).

Another indication of the improvements offered by A/R+ is the relationship between the interval $[\hat{p}_*(X_q|\mathbf{X}_E), \hat{p}^*(X_q|\mathbf{X}_E)]$ generated by A/R+ and the same interval generated by A/R. This is illustrated in Table 2, using the same graph in Figure 1, but now computing the probability intervals for the event $\{L = l_0\}$. Now we generated 100 different credal networks with the same underlying graph, and took the average of the interval lengths. Note the decrease in interval length (remember that the correct intervals are enclosed by the intervals

Table 2: Amplitude of probability intervals for $\{L = l_0\}$, for graph in Figure 1. Tests with four different combinations of number of categories and number of vertices; ensembles of 100 networks.

| number of categories | number of vertices | A/R mean interval length | A/R+ mean interval length |
|---|---|---|---|
| 03 | 02 | 0.47 | 0.39 |
| 03 | 03 | 0.61 | 0.56 |
| 03 | 04 | 0.67 | 0.63 |
| 04 | 02 | 0.45 | 0.36 |
| 04 | 03 | 0.52 | 0.47 |

generated by A/R+).

To avoid the possibility that differences between A/R+ and A/R are just being magnified by the particular graph in Figure 1, we considered a large set of randomly generated polytrees (using the generator described in [19]). In all inferences we could try, we verified the same pattern of reduction in the relative error. We generated 9 different polytrees, each with 20 variables; for each polytree, we generated 5 sets of local credal sets. For each node of this set of networks, we run the A/R and the A/R+ algorithms, and in some of them we compared the approximations with the exact inferences (obtained with the branch-and-bound algorithms described in Section 7). The mean length of the intervals generated by the A/R algorithm are significantly larger than the mean length for the A/R+ algorithm. More importantly, the A/R+ algorithm produces bounds that are about an order of magnitude more precise (in relative error) than the bounds generated by the A/R algorithm. We return to such results at the end of Section 6 and in Section 7.

In closing, we note that the complexity of the set-valued messages in A/R+ can become unmanageably large in some situations. When we must compute a message that requires an excessively complex credal set, we simply resort to the original approximation proposed by Tessem (we have a threshold indicating the maximum number of vertices the algorithm should handle explicitly).[4]

## 6  Inner approximation: Multilinear programming

An inner approximation for $\bar{p}(X_q|\mathbf{X}_E)$ can be generated by any method that looks for a local maxima of $p(X_q|\mathbf{X}_E)$ subject to constraints imposed by local credal sets $K(X_i|\text{pa}(X_i))$. Such approximations have been considered in previous literature, using for example simulated annealing [4] and genetic algorithms [5]. Generally these methods require tuning several parameters; we have implemented some of them and noticed that, while they produce reasonable solutions, they are far from easy to apply. Other ideas, such as gradient search or geometric programming, have also been proposed but not implemented so far [1, 9, 28] — such techniques would demand a great deal of numerical finesse and would have to be carefully adapted to credal networks.

---

[4]The A/R+ algorithm can be made even more flexible: we could try to pass the credal sets $\pi_{Z_i}(X)$ as messages whenever possible (without applying the re-conversion to interval-valued form), until we reach a point where the number of vertices in the messages exceeds some limit. We have not tried this option so far.

The computation of $\bar{p}(X_q|\mathbf{X}_E)$ is a maximization problem with constraints on $K(X_i|\text{pa}(X_i))$ and an objective function that is a ratio of two large multilinear functions represented by (1) (such a problem is usually classified as a *signomial* program [2]). In this section we ask, can we exploit the fact that, in our problem, 1) all constraints are linear; 2) all constraints are in some sense "local" (they are grouped with the local credal sets $K(X_i|\text{pa}(X_i))$); and 3) multilinear functions are possibly the simplest signomial functions to be found? We now present a new algorithm that does benefit from these properties, with no free parameters to tune, and not requiring special methods to control numerical error.

We are inspired by Lukatskii-Shapot's algorithm for local search in multilinear programs [22]. Lukatskii and Shapot assume that a multilinear function must be optimized subject to linear constraints. The Lukatskii-Shapot algorithm is simple: fix every variable in the multilinear problem, except one, and solve the resulting linear program to optimality: then fix another variable, and so on, until no improving change is possible. The algorithm surely terminates because a multilinear program has its minima/maxima in the vertices of the feasible region.

Our inferences are not exactly multilinear programs (the objective function is a *fraction* of multilinear functions), but inferences do keep an essential property used by the Lukatskii-Shapot algorithm: every minima/maxima of $p(X_q|\mathbf{X}_E)$ in a strong extension must occur at a vertex of the extension [17]. That is, we can fix a vertex for every credal set except one, and just check which vertex of the remaining credal set minimizes/maximizes $p(X_q|\mathbf{X}_E)$ (given that all the others are fixed). We retain the minimizing/maximizing vertex, and then move to the next credal set. We now fix all the vertices except for this next credal set, *using the minimizing/maximizing* vertex obtained in the previous step. We keep repeating this, going over and over all the local credal sets in the credal network. The process is surely to stop: every step increases the objective function, and there is only a finite number of possible moves (given that variables are discrete and local credal sets have finitely many vertices).

The local search procedure just described can be easily implemented as follows. Assume the search aims at the upper probability $\bar{p}(X_q = x_q|\mathbf{X}_E)$. Note the use of additional variables and a MAP algorithm to organize the process:

1. Select an ordering for the nodes in the credal network.

2. Select a vertex for every local credal set, and compute $p(X_q = x_q|\mathbf{X}_E)$ using these vertices (this is the initial value).

3. Repeat until there is no change on the value of $p(X_q = x_q|\mathbf{X}_E)$, cycling over the vertices in the ordering (the current vertex is denoted by $V$):

   (a) Keep all vertices fixed except for a credal set $K(V|\text{pa}(X))$ associated with node $V$.

   (b) Add a "transparent" variable $V'$ as an additional parent of $V$; the transparent variable $V'$ has as many categories as the number of vertices in $K(V|\text{pa}(V))$, and now $p(V|V', \text{pa}(V))$ aggregates all vertices of $K(V|\text{pa}(V))$.

   (c) Run a MAP algorithm in the transformed network, where $V'$ is the only maximizing variable.

   (d) The best value of $V'$ obtained by MAP indicates the vertex of $K(V|\text{pa}(V))$ that locally maximizes $p(X_q = x_q|\mathbf{X}_E)$; fix $p(V|\text{pa}(V))$ at this vertex.

4. Stop when there is no change and return the current $p(X_q = x_q|\mathbf{X}_E)$.

The algorithm can be easily modified to approximate lower probabilities. Note that this algorithm uses "transparent" variables as suggested by Cano et al [4], to process a credal network as a standard Bayesian network. Note also that the algorithm is completely general (not restricted to polytrees), but it is particularly efficient in polytrees. Finally, note that the ordering of nodes generated in the first step need not be arbitrary; in fact some orderings may be better than others (we thank one reviewer for this observation).

More importantly, the local search algorithm just proposed typically produces very accurate approximations. We have run it in a large number of medium-sized networks, and verified that in most cases it finds the exact answer (we have resorted to the branch-and-bound algorithm in Section 7 to compute exact values). A mean comparison with exact inferences is perhaps not very illuminating, as it is not possible to find exact answers for the largest (and more interesting) networks; consider instead the following experiment. We took the same random networks described in Section 5 and considered the computation of upper probabilities for all variables in the networks. We then produced *outer* approximations with the A/R and A/R+ algorithms, and *inner* approximations with the search-based algorithm. These approximations are quickly produced, and we know that the exact value of upper probabilities is between the inner and the outer approximations. Denote by $B1$ the difference between the A/R outer bound and the search-based inner bound for an upper probability. Denote by $B2$ the difference between the A/R+ outer bound and the search-based inner bound for an upper probability. Now, how precise are $B1$ and $B2$? We have found that intervals from $B2$ are usually an order of magnitude smaller than intervals from $B1$. A more remarkable point is that the length of intervals from $B2$ is very small, thereby mutually corroborating the power of A/R+ and the search-based method.

The experiments in the previous paragraph suggest the following scenario: with relatively little effort, one can generate an outer approximation with A/R+ and an inner approximation with the search-based algorithm. It is likely that most applications will find these approximations to "bracket" the exact answer to a satisfactory level. In case more accuracy is needed, the methods in the next section must be used.

## 7 Branch-and-bound search for strong extension

We have so far produced two methods for bounding lower/upper probabilities. In this section we describe strategies that combine these bounds to search for exact inferences, using branch-and-bound procedures [3, 24]. To our knowledge, these procedures are the first explicit formulation and implementation of inference in credal networks as a search procedure that runs to optimality. The presentation is brief and more details about the general method can be found in [13]; in this paper we are interested in the behavior of branch-and-bound algorithms that use the A/R+ algorithm and the inner approximations presented in previous sections.

In this section we focus only on the computation of upper probabilities; the computation of lower probabilities is completely analogous.

Given a maximization problem $P$, a branch-and-bound algorithm divides $P$ in sub-instances that are easier to solve or approximate than $P$. The partitioning is made so that the solution for $P$ is present in one of the sub-instances [23]. Each sub-instance, $R$, is evaluated with a relaxed algorithm that produces a bound $r(R)$ (overestimation). This process is repeated for each sub-instance while there is a promising alternative. A *depth-first* branch-and-bound follows the most promising sub-instance as soon as possible, while pruning sub-instances that are guaranteed not to contain the maximum. A *breadth-first* branch-and-bound stores the whole "frontier" of sub-instances.

In our context, a sub-instance of a credal network is generated by fixing one vertex of a local credal set. If we were to organize the branch-and-bound procedure as a search tree, we would see the complete network at the root, with a gradual "thinning" of the local credal sets — the leaves are standard Bayesian networks. Outer bounds (from A/R+) are

Table 3: Cost of exact inference for $\bar{p}(E = e_0)$ in the graph of Figure 1 with depth-first branch-and-bound. The cost is measured as the number of nodes expanded in a depth-first branch-and-bound search. Ensembles of 35, 10 and 10 networks.

| Case | Time(s) | Cost (mean) | Cost (deviation) |
|---|---|---|---|
| With A/R | | | |
| a | 5.4 | 3356.8 | 1314.1 |
| b | 395 | 254559.3 | 139214.1 |
| c | 1511 | 527756.8 | 187673.9 |
| With A/R+ | | | |
| a | 0.97 | 365.7 | 394.8 |
| b | 17.44 | 7271.2 | 9338.6 |
| c | 584 | 37143.7 | 37576 |

(a) 3 categories for variable/2 vertices by credal set
Complexity of exhaustive algorithm = $2^{21}$ vertices
(b) 3 categories for variable/3 vertices by credal set
Complexity of exhaustive algorithm = $3^{21}$ vertices
(c) 4 categories for variable/2 vertices by credal set
Complexity of exhaustive algorithm = $2^{35}$ vertices used to overestimate sub-instances and direct the search, while inner bounds (from the search-based method) are used to prune sub-instances. When a leaf is reached, we can run standard inference and obtain the exact probability value.

The advantage of depth-first branch-and-bound is the minimal memory consumption. The advantage of breadth-first branch-and-bound is that it allows us to gradually obtain improving approximations; we can gradually refine outer and inner bounds by looking at all the nodes in the "frontier". The disadvantage of breadth-first is the potentially enormous cost in memory. We only describe tests with depth-first branch-and-bound, as our main interest in this section is to discuss exact inference.

We ran experiments with networks containing variables with tree and four states. Each configuration was tested again using several randomly generated credal nets [19].

First take the polytree in Figure 1. Results for queries on variable $E$ (with depth-first branch-and-bound) are reported in Table 3. The table shows results when the branch-and-bound algorithm uses A/R and A/R+ as bounding methods.

We observe that the size of the search tree explored by branch-and-bound is usually a tiny fraction of the potential number of vertices of the strong extension. Note the enormous difference between potential vertices of the strong extension and actually expanded vertices. We also can see that A/R+ is superior to A/R.

As another example of the efficiency of the algorithm, take the computation of $\bar{p}(H = h_0)$ in the graph of Figure 1, with variables containing three categories, and with a random collection of credal sets, where each credal set has three vertices. In this case, depth-first branch-and-bound obtained the exact solution after examining just 4634 vertices of the strong extension — note that the strong extension potentially contains $3^{50}$ vertices. The relative error between the exact result and the inner bound, and the exact result against A/R+ are 0.002 and 0.015 respectively.

We have observed such behavior in many experiments on randomly generated networks. We have observed that polytrees with up to 10 variables can be usually handled without problems. In closing, we comment that networks with the graph in (2) can be solved almost instantly by depth-first branch-and-bound. Consequently, a network with mostly binary variables but with a fragment such as (2) would be a case where previous algorithms would fail, while the present methods would easily produce a solution.

## 8 Conclusion

In this paper we have presented three new ideas concerning inference in credal networks:

1. The A/R+ algorithm, a significant improvement on Tessem's A/R algorithm for the outer approximations.

2. A local search algorithm, inspired by the Lukatskii-Shapot algorithm, that produces accurate inner approximations.

3. Branch-and-bound algorithms that use the two previous techniques to produce exact/approximate solutions.

The A/R+ algorithm and the local search algorithm can be used together to produce an enclosing interval containing the lower/upper probability of interest. Typically this interval is rather small and should be useful in many applications. Branch-and-bound methods can then be used to refine these intervals or to actually produce exact solutions. It should be noted that the branch-and-bound algorithms described here can be best understood as a *family* of methods for exact and approximated inference in strong extensions. Working separately or together, these techniques offer improvements on the efficiency of inferences in credal networks — often the computational gains are on many orders of magnitude.

It seems reasonable to expect that credal networks will be processed by approximate algorithms in most cases; however, today it is critical to have exact algorithms that can be used to test approximations and to provide "ground truth" in experiments. At the current stage, it would seem that exact algorithms with the ability to handle medium-size networks (say up to 15 to 30 non-binary variables) would be powerful enough to produce such "ground-truth." The purpose of branch-and-bound techniques we presented was to move in the direction of this goal for polytrees.

Although we have restricted ourselves to polytrees, branch-and-bound techniques can be used for general inference provided that bounds are available. Generalizations of Tessem's bounds are possible (following Ha et al [18]), and the same combination of A/R and separable variable elimination can be used in those methods. Likewise, the local search method can be readily applied to general networks with minor modifications.

We also would like to emphasize the possibility that a network be processed in pieces, using different levels of accuracy in each one of the partial inferences. Such a strategy seems to be appropriate for large networks. Future research will be focused on developing better bounds and heuristics for search, and on experimenting variations of the branch-and-bound scheme — including the possibility of processing networks in pieces.

## Acknowledgements

We thank Jaime Ide for generating random networks that were used in our tests.

This work has received generous support from HP Labs; we thank Marsha Duro from HP Labs for establishing this support and Edson Nery from HP Brazil for managing it. The work has also been supported by CNPq (through grant 3000183/98-4) and by CAPES.